\documentclass{article}
\usepackage{graphicx}

     \PassOptionsToPackage{numbers, compress}{natbib}
\usepackage[preprint]{neurips_2022}

\usepackage[utf8]{inputenc} % allow utf-8 input
\usepackage[T1]{fontenc}    % use 8-bit T1 fonts
\usepackage{hyperref}       % hyperlinks
\usepackage{url}            % simple URL typesetting
\usepackage{booktabs}       % professional-quality tables
\usepackage{amsfonts}       % blackboard math symbols
\usepackage{nicefrac}       % compact symbols for 1/2, etc.
\usepackage{microtype}      % microtypography
\usepackage{xcolor}         % colors
\usepackage{amsmath}
\usepackage{bm}      

\title{Deformable Capsules  for Object Detection}

\author{%
  Rodney LaLonde\\
Palantir Technologies, Washington, DC. \\
  \texttt{lalonderodney@gmail.com } \\
  \And
  Naji Khosravan\thanks{Equal contribution to the first author} \\
  Zillow Group Inc., Seattle, WA  \\
  \texttt{najik@zillowgroup.com} \\
  \AND
  Ulas Bagci \\
  Dept of ECE, BME, and Radiology\\
  Northwestern University,  Chicago, IL, 60611 \\
  \texttt{ulasbagci@gmail.com} \\
}

\begin{document}

\maketitle

\begin{abstract}
 Capsule networks promise significant benefits over convolutional networks by storing stronger internal representations, and routing information based on the agreement between intermediate representations' projections. Despite this, their success has been limited to small-scale classification datasets due to their computationally expensive nature. Though memory efficient, convolutional capsules impose geometric constraints that fundamentally limit the ability of capsules to model the pose/deformation of objects. Further, they do not address the bigger memory concern of class-capsules scaling up to bigger tasks such as detection or large-scale classification. In this study, we introduce a new family of capsule networks, deformable capsules (\textit{DeformCaps}), to address a very important problem in computer vision: object detection. We propose two new algorithms associated with our \textit{DeformCaps}: a novel capsule structure (\textit{SplitCaps}), and a novel dynamic routing algorithm (\textit{SE-Routing}), which balance computational efficiency with the need for modeling a large number of objects and classes, which have never been achieved with capsule networks before. We demonstrate that the proposed methods efficiently scale up to create the first-ever capsule network for object detection in the literature. Our proposed architecture is a one-stage detection framework and it obtains results on MS COCO which are on par with state-of-the-art one-stage CNN-based methods, while producing fewer false positive detection, generalizing to unusual poses/viewpoints of objects.
\end{abstract}

\section{Introduction}
Capsule networks promise many potential benefits over convolutional neural networks (CNNs). These include practical benefits, such as requiring less data for training or better handling unbalanced class distributions~\cite{jimenez2018capsule}. There are  important theoretical benefits in using capsules such as building-in stronger internal representations of objects~\cite{punjabi2020examining}, and modeling the agreement between those intermediate representations which combine to form final object representations (e.g. part-whole relationships)~\cite{kosiorek2019stacked,sabour2017dynamic}. Although these benefits might not be seen yet in the performance metrics (e.g. average precision) on standard benchmark computer vision datasets, they are important for real-world applications. As an example, it was found by~\cite{alcorn2019strike} that \textit{CNNs fail to recognize 97\% of their pose space}, while capsule networks have been shown to be far more robust to pose variations of objects~\cite{hinton2018matrix}; 
%,lalonde2020capsules};
further, real-world datasets are not often as extensive and cleanly distributed as ImageNet or MS COCO.

These benefits are achieved in capsule networks by storing richer vector (or matrix) representations of features, rather than the simple scalars of CNNs, and dynamically choosing how to route that information through the network. The instantiation parameters for a feature are stored in these capsule vectors and contain information (e.g. pose, deformation, hue, texture) useful for constructing the object being modeled. Early studies have shown strong evidence that these vectors do in fact capture important local and global variations across objects' feature components (or parts) within a class~\cite{punjabi2020examining,sabour2017dynamic}. Inside their networks, capsules dynamically route their information, seeking to maximize the agreement between these vector feature representations and the higher-level feature vectors they are attempting to form.

Despite their potential benefits, many have remained unconvinced about the general applicability of capsule networks to large-scale computer vision tasks. To date, no capsule-based study has achieved classification performance comparable to a CNN on datasets such as ImageNet, instead relegated to smaller datasets such as MNIST or CIFAR. Worse still, to the best of our knowledge, \textit{no capsule network has shown successful results in object detection}, a very important problem in computer vision, robotics, and medical imaging. Now, the argument can be made that standard benchmark datasets such as ImageNet or MS COCO likely contain majority of objects in that $3\%$ range of usual poses, and thus CNNs will appear to perform extremely well when measured in terms of accuracy, stripping capsule networks of one of their largest advantages. However, until capsule networks can perform on-par with CNNs on these typical object poses, few will care about the benefits of stronger internal representations and better generalization to unseen poses.

\textbf{Summary of Our Contributions:} (1) We propose the first ever capsule-based object detection framework in the literature. Our network is a one-stage (single-shot) architecture, where objects are both localized and classified using capsules, and can perform on-par with the state-of-the-art CNNs on a large-scale dataset (MS COCO). (2) We address the geometric constraint of convolutional capsules (and locally-constrained routing) by introducing \textit{deformable capsules}, where parent capsules learn to adaptively sample child capsules, effectively eliminating rigid spatial restrictions while remaining memory efficient. (3) We design a new capsule-based prediction head structure, \textit{SplitCaps}, which reformulates the projections of an objects’ instantiation parameters, presence, and class, eliminating the previous dimensional increase of capsules by the number of classes. This crucial addition enables the training of capsule networks on large-scale computer vision datasets for the first time in the literature. (4) To route information across \textit{SplitCaps'} unique structure, we introduce a novel Squeeze-and-Excitation inspired dynamic routing algorithm, \textit{SE-Routing}, which seeks to maximize agreement between child capsule projections, without the need of iterative loops.

\begin{figure*}[t]
\centering
\includegraphics[scale=0.6]{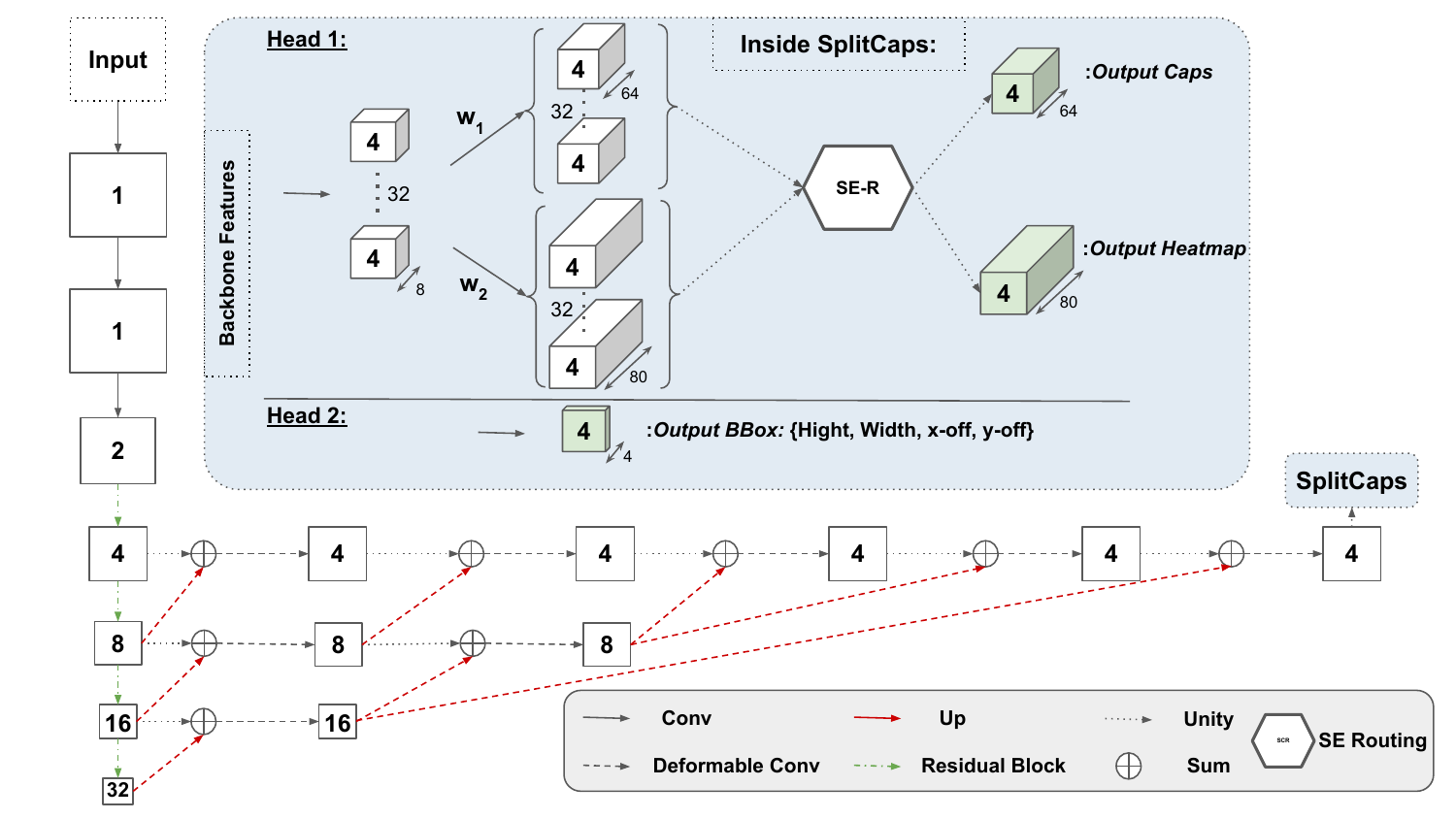}
\caption{Deformable capsule architecture for object detection.} \label{fig:arch}
\end{figure*}

\section{Deformable capsules: Fixing locally-constrained dynamic routing} \label{sec:deformcaps}

The capsule network architecture proposed by~\cite{sabour2017dynamic} acted on global information, where digit capsules represented the pose and presence of digits in an image regardless of spatial location. The information from all children in the previous layer was sent to every parent in the following layer, weighted via the routing coefficients found in a cosine similarity routing algorithm. While this proved to be a highly-effective strategy, it was also computationally expensive, limiting its use to \textit{only} small-scale datasets. Recent works attempted to scale up capsule networks to larger problems such as biomedical image segmentation~\cite{lalonde2018capsules} and classification~\cite{lalonde2018capsules,lalonde2020diagnosing} or action detection in video~\cite{duarte2018videocapsulenet} by using convolutional capsules and locally-constraining the routing algorithm. Although efficient solutions were presented in those studies, the representation power of capsule networks was fundamentally limited due to imposing local constraints. This is because convolutions, by design, have a fixed geometric structure~\cite{dai2017deformable}, and such a geometric constraint significantly inhibits capsules' ability to model part-whole relationships, relying on parts of objects to fall within a fixed local grid. Therefore, it is unreasonable to expect a capsule to effectively represent the pose and deformations of an object when the information related to the parts of that object are locked into a fixed spatial relationship. Locally-constrained routing was a needed step to make capsules viable on larger-scale problems, but in some ways it was fundamentally a step in the wrong direction.

In this study, we effectively solve this aforementioned problem by introducing a method that balances efficiency with the ability for a capsule to represent any pose and deformation of an object (i.e. where child capsules can be found in different spatial relationships to one another for the same parent). In such a formulation, global information is not explicitly required, but it does require parent capsules to have more flexibility over which child capsules they draw information from. Our proposed solution is \textit{deformable capsules}. The idea behind the proposed algorithm is simple: if parent capsules are supposed to capture common deformations of the objects they represent within their vectors, then the choice of which children to aggregate information from must be handled in a deformable manner as well. Deformable capsules allow parents to adaptively gather projections from a non-spatially-fixed set of children, and thus \textit{effectively and efficiently} model objects' poses.

To achieve this overall goal, we follow the same efficient convolutional capsule paradigm, where projection vectors are formed via a convolution operation with a kernel centered on the parent capsules' spatial location, but now we learn an additional set of weights for each parent capsule. These learnable weights are the same shape as each parent's kernel, and represent the offset values for the spatial sampling of child capsules for that parent. Based on the child capsule representation vectors in the previous layer, these weights learn which children a parent capsule should adaptively sample from for a given input image. Dynamic routing then determines how to weight the information coming from each of these children based on their agreement for each projected parent.

A concrete example to better illustrate the parameter savings of the proposed technique is given in the following section. In summary, we found deformable capsules to converge faster and to much higher performance than convolutional capsules.

%%%%%%%%%%%%%%%%%%%%%%%%%%%%%%%%%%%%%%%%%%%%%%%%%%%%%%%%
\section{Objects as capsules: \textit{SplitCaps} with \textit{SE-Routing}} \label{sec:objects_as_caps}
We propose a novel one-stage (single-shot) capsule network architecture for object detection, called \textit{DeformCaps}, where objects are detected, classified, and modeled with capsules. Our overall network architecture, shown in Fig.~\ref{fig:arch}, is built upon \textit{CenterNet} by \cite{zhou2019objects} who proposed to represent objects in images as scalar point values located at the center of their bounding boxes. The authors then regress the remaining characteristics of the object (e.g. height, width, depth) for each center-point detected. In our work, we follow the same center-point detection paradigm, but represent our objects with capsule vectors instead. Since several recent studies have found utilizing a CNN backbone before forming capsule types to be beneficial to overall performance~\cite{duarte2018videocapsulenet,kosiorek2019stacked,tsai2020capsules}, we adopt the preferred backbone of CenterNet, DLA-34~\cite{yu2018deep}, as this gave the best trade-off between speed and accuracy. Features extracted from the backbone are sent to our capsule object detection head and a bounding box regression head. Then, we introduce a new capsule structure, called \textit{SplitCaps}, composed of class-agnostic and class-presence capsules, and a novel routing algorithm, called \textit{SE-Routing} to detect objects with capsules.

%%%%%%%%%%%%%%%%%%%%%%%%%%%%%%%%%%%%%%%%%%%%%%%%%%%%%%%%%%%%%%%%%%%%%%%%%%%
\subsection{SplitCaps: Class-agnostic and class-presence capsules} \label{sec:splitcaps}
%%%%%%%%%%%%%%%%%%%%%%%%%%%%%%%%%%%%%%%%%%%%%%%%%%%%%%%%%%%%%%%%%%%%%%%%%%%

The original CapsNet by~\cite{sabour2017dynamic} was extremely expensive in computation and our proposed deformable capsules is a best possible solution to balance non-rigid deformations of objects while remaining memory efficient. However, \textit{there is a more significant memory hurdle capsule networks must overcome when scaling up to large-scale datasets}, such as MS COCO. In their current implementation, capsule networks represent each class with its own parent capsule vector. On small scale classification datasets (and using deformable routing), this is not an issue; it amounts to $2 \times k \times k \times a_i \times c_j \times a_j$ parameters and $N \times c_i \times c_j \times a_j \times 4$ bytes to store the intermediate representations to be routed for the parents, where $c_j$ is usually around $10$ classes to represent. Let us suppose we have $5 \times 5$ kernels with $32$ input capsule types of $8$ atoms per capsule, $10$ output capsule types of $16$ atoms per capsule, and a batch size of $32$. In total, we would have $2 \times 5 \times 5 \times 8 \times 10 \times 16 = 64$K parameters and $32 \times 32 \times 10 \times 16 \times 4 \approx 655$ KB.

When we scale this up to object detection, we now need to store representations \textit{for every possible object location}, and for MS COCO, for instance, we need to \textit{represent 80 possible classes}. This gives us $2 \times k \times k \times a_i \times c_j \times a_j \Rightarrow 2 \times 5 \times 5 \times 8 \times 80 \times 16 = 512$K parameters and $N \times H \times W \times c_i \times c_j \times a_j \times 4$ bytes $\Rightarrow 32 \times 128 \times 128 \times 32 \times 80 \times 16 \times 4 \approx 86$ GB for the intermediate representations, where we assume the output grid of detections is $128 \times 128$ with a single detection (i.e. bounding box) predicted per class per location. The problem is not any better for large-scale classification datasets such as ImageNet either, where we lose the grid of predictions but grow to $1000$ classes, which would require $2 \times 5 \times 5 \times 8 \times 1000 \times 16 = 6.4$M parameters and $32 \times 32 \times 1000 \times 16 \times 4 \approx 66$ GB for the intermediate representations. Clearly, with most GPU memories limited to $12$--$24$ GB, a solution is needed to be found for capsule networks to scale up to larger-scale computer vision tasks.

To overcome this issue, we propose a new type of capsule architecture, \textit{SplitCaps}, to more efficiently scale capsule networks to large-scale computer vision tasks. SplitCaps contains two parent capsule types, each with a different number of atoms per capsule, for each location of the detection grid. As before, the idea is to balance efficiency with the ability to learn powerful representations. Towards this goal, SplitCaps proposes to divide up between its two parent capsules the tasks of \textit{(i)} learning the instantiation parameters necessary to model the possible variations of an object and \textit{(ii)} predicting which classes of objects are present in a given input. The first capsule type we refer is our class-agnostic object instantiation capsules, and the second we refer is class presence capsules.

%%%%%%%%%%%%%%%%%%%%%%%%%%%%%%%%%%%%%%%%%%%%%%%%%%%%%%
\textbf{Class-agnostic object instantiation capsules:} 
%%%%%%%%%%%%%%%%%%%%%%%%%%%%%%%%%%%%%%%%%%%%%%%%%%%%%% Don't add space below this line here!
The purpose of these capsules is similar to those in previous works: model the possible variations (in pose, deformation, texture, etc.) of objects within a vector of instantiation parameters, the span of which should cover all possible variations for that object at test (hence why capsules are better than CNNs at generalizing to unseen poses). While previous capsule networks did this in a class-wise manner, we argue such a formulation is not required and possibly it is redundant. Many variations (e.g. rotation, skew, stroke thickness) may be class-independent, and thus to model these variations class-wise would require repetition across each capsule type. Instead, we propose to model all classes within a single capsule type (i.e. class-agnostic). In this way, while class-dependent variations would each require their own dimensions of the vector, class-independent variations can each be modeled along a single dimension for all possible objects. Since it is reasonable to assume there will be at least some class-specific variations, we increase the default capsule vector dimension from $16$ to $64$ to accommodate for possible class-specific instantiation parameters.

During training, if an object is present at a given spatial location, the 64-dimensional capsule vector for that location is fed to a reconstruction regularization sub-network to construct the mask of that object as similar to the reconstruction regularization used by~\cite{sabour2017dynamic} for classification. This sub-network is a relatively small and fast addition: a set of three ReLU-activated $1 \times 1$ convolutional layers with $256$ filters each, followed by a final sigmoid-activated $1 \times 1$ convolution with $N = n^2 = 28^2 = 784$ filters, before reshaping outputs to $n \times n$. Since objects' scales vary dramatically, we scale normalize all objects' ground-truth masks to be $28 \times 28$ (by following~\cite{he2017mask}). Supervised training is then conducted by computing the Dice loss~\cite{milletari2016v} between the predicted reconstruction, $r$, and the object's mask, $m$
\begin{equation}
   \mathcal{L}_r = \frac{2\sum_{i}^{N}r_{i}m_{i}}{\sum_{i}^{N}r_{i}^{2}+\sum_{i}^{N}m_{i}^{2}} \textrm{,}
\end{equation}
where $\mathcal{L}_r$ is used to provide a regularization signal to the instantiation parameters being learned.

%%%%%%%%%%%%%%%%%%%%%%%%%%%%%%%%%
\textbf{Class presence capsules:}
%%%%%%%%%%%%%%%%%%%%%%%%%%%%%%%%% Don't add space below this line here!
The class presence capsules attempt to model which classes of objects are present in the input at each spatial location, if any. We can accomplish this by setting the atoms per capsule to the number of classes being represented (i.e. 80 for MS COCO). Just as~\cite{hinton2018matrix} separately modeled pose (with a matrix) and activation (with a scalar), this 80-dimensional vector can be viewed as a class-dependent set of activation values. The activation values are then passed through a sigmoid function and thresholded; if one or more activation values are above the threshold, an object is determined to be at that spatial location with the strongest activated dimension determining the class label.

In order to produce a smooth loss function during training, we create a ground-truth heatmap by fitting a Gaussian distribution rather than a single point, to the center-point of each object's bounding box, with variance proportional to the size of the box following~\cite{zhou2019objects}. More specifically, we create a heatmap ${H \in [0,1]^{\frac{X}{d} \times \frac{Y}{d} \times K}}$ containing each down-scaled ground truth center-point $\tilde p = \big(\frac{p_x}{d},\frac{p_y}{d} \big)$ for class $k \in K$ using a Gaussian kernel ${H_{xyk} = \exp\left(-\frac{(x-\tilde p_x)^2+(y-\tilde p_y)^2}{2\sigma_p^2}\right)}$, where $d$ is the amount of downsampling in the network and $\sigma_p$ is an object-size-adaptive standard deviation~\cite{law2018cornernet}. In the case of overlapping Gaussians, we take the element-wise maximum. To handle the large class imbalance between objects and background in our heatmaps, we use a penalty-reduced pixel-wise logistic regression with a focal loss~\cite{lin2017focal}:

\begin{equation}
    \mathcal{L}_h = \frac{-1}{P} \sum_{xyk}
    \begin{cases}
        (1 - \hat{H}_{xyk})^{\alpha} 
        \log(\hat{H}_{xyk}) \&  \! \text{ if }\ H_{xyk}=1,\vspace{2mm}\\
        (1-H_{xyk})^{\beta} (\hat{H}_{xyk})^{\alpha} \log(1-\hat{H}_{xyk}),
         \& \! \text{ otherwise; }
    \end{cases}
\end{equation}

where $\alpha, \beta$ are hyper-parameters of the focal loss, $P$ is the number of center-points in the input, used to normalize all positive focal loss instances to $1$~\cite{zhou2019objects}. We use $\alpha=2$ and $\beta=4$ in all our experiments, following~\cite{law2018cornernet}. At test, to efficiently retrieve the object's exact center, we run a $3 \times 3$ max-pooling over the thresholded spatial map. 

To predict the height and width of the bounding boxes of objects and recover the $x,y$ offsets needed to map back to the upscaled image dimensions, we follow the same formulation as~\cite{zhou2019objects} and pass the backbone features through a $3 \times 3$ convolutional layer with $256$ feature maps, then a $1 \times 1$ convolutional layer with $2$ feature maps. These layers predict the local offset, $\hat O \in \mathbb{R}^{\frac{W}{d} \times \frac{H}{d} \times 2}$, and size prediction, $\hat S \in \mathbb{R}^{\frac{W}{d} \times \frac{H}{d} \times 2}$, for each center-point and are supervised by 
\begin{equation}
    \mathcal{L}_{o} = \frac{1}{P}\sum_{p} \left|\hat O_{\tilde p} - \left(\frac{p}{d} - \tilde p\right)\right|,
\end{equation}
and
\begin{equation}
\quad \mathcal{L}_{s} = \frac{1}{P}\sum_p \left|\hat S_{p} - (x_2 - x_1, y_2 - y_1)\right|,
\end{equation}
\normalsize
respectively. Our final objective function is thus defined as $\mathcal{L} = \mathcal{L}_{h} + \lambda_{r} \mathcal{L}_{r} + \lambda_{s} \mathcal{L}_{s} + \lambda_{o} \mathcal{L}_{o}$. We keep $\lambda_{s} = 0.1$ and $\lambda_{o} = 1$ as done in~\cite{zhou2019objects}, and set $\lambda_r = 0.1$ initially, then step up to $\lambda_r = 2.0$ at the half-way in training.

\begin{figure*}
\centering
\includegraphics[scale=0.5]{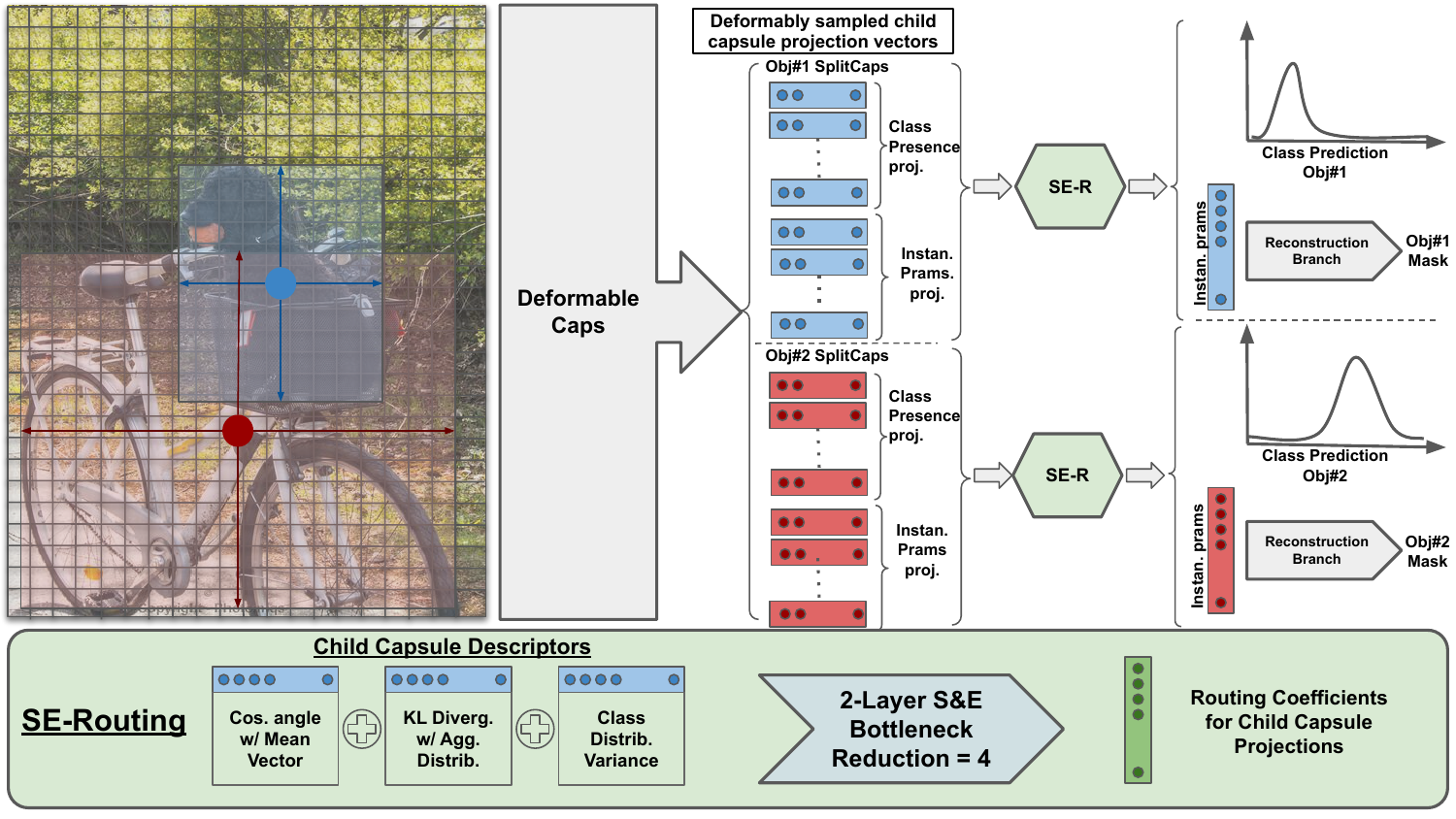}
\caption{Proposed deformable SplitCaps formulation which, in one-shot, localizes all objects to their centers, determines their classes, and models their instantiation parameters in two parallel parent capsule types. Information is dynamically routed from children to parents by passing three chosen child capsule descriptors through a two-layer Squeeze-and-Excitation (S\&E) bottleneck.} \label{fig:routing}
\end{figure*}

%%%%%%%%%%%%%%%%%%%%%%%%%%%%%%%%%%%%%%%%%%%%%%%%%%%%%%%%
\subsection{SE-Routing: SplitCaps new routing algorithm} \label{sec:se_routing}
%%%%%%%%%%%%%%%%%%%%%%%%%%%%%%%%%%%%%%%%%%%%%%%%%%%%%%%%
Since \textit{SplitCaps} introduces a unique capsule head structure, where instantiation parameters and activations are split across different capsule types, previous dynamic routing algorithms can no longer be directly applied. To overcome this, we propose a new dynamic routing algorithm that takes inspiration from Squeeze-and-Excitation networks~\cite{hu2018squeeze}, which we call \textit{SE-Routing}, and illustrated in Fig.~\ref{fig:routing}. 

\medskip

Previously proposed dynamic routing algorithms (e.g.~\cite{sabour2017dynamic} and~\cite{hinton2018matrix}) were typically iterative, requiring a hand-tuned loop of routing iterations, which proved to be slow and temperamental in practice. Different studies found different numbers of iterations to be effective, and one meta-study of five different iterative dynamic routing algorithms found them all to be largely ineffective~\cite{paik2019capsule}. To avoid this pitfall, we propose a new routing algorithm to dynamically assign weights to child capsule projections based on their agreement, computed in a single forward pass using a simple gating mechanism with sigmoid activation. Unlike~\cite{sabour2017dynamic} which uses a routing softmax to force a one-hot mapping of information from each child to parents, our proposed SE-Routing learns a non-mutually-exclusive relationship between children and parents to allow multiple children to be emphasised for each parent. 

%\medskip
%%%%%%%%%%%%%%%%%%%%%%%%%%%%%%%%%%%%%%%%%%%%%%%%%%%%%%%%%%%%%%%%%
\subsubsection{Creating child capsule projection descriptors (squeeze):}
%%%%%%%%%%%%%%%%%%%%%%%%%%%%%%%%%%%%%%%%%%%%%%%%%%%%%%%%%%%%%%%%% Don't add space below this line here!
Following the Squeeze-and-Excitation paradigm, we first must compute the squeeze (i.e. a set of descriptors which summarize relevant information about each feature) to create a set of child capsule projection descriptors. In~\cite{hu2018squeeze}, the authors proposed to use the global average activation of each channel with the goal of modeling channel inter-dependencies. In this study, our goal is to maximize the agreement between child projections, for both the instantiation parameters and class presence of the object being modeled. With that motivation, we compute three separate descriptors which are fed into the excitation phase of the routing: \textit{(i)} cosine angle between the mean projection vector and each child’s projection, which captures \textit{object instantiation agreement}; \textit{(ii)} Kullback–Leibler (KL) divergence of each child’s predicted class distribution and an aggregated distribution, which captures \textit{class presence agreement}; and \textit{(iii)} variance of each child’s predicted class distribution, which captures \textit{class presence uncertainty}.

\medskip

The cosine angle descriptor, $\bm{a}$, is calculated in a similar manner to~\cite{sabour2017dynamic}. A mean projection vector, $\bm{\tilde u} = 1/N\sum_i^N \bm{\hat u}_i$, is first computed using the set of child capsule projections, $\bm{\hat U} = \{\bm{\hat u}_1, \bm{\hat u}_2, ..., \bm{\hat u}_N\}$. Then, we compute a set of cosine angles between each individual projection and this mean, $\bm{a} = \{a_1, a_2, ..., a_N\}$, where $a_i = (\bm{\tilde u} \cdot \bm{\hat u}_i) / (\lvert \bm{\tilde u} \rvert \cdot \lvert \bm{\hat u}_i \rvert)$.

\medskip

In a similar fashion, we compute a KL divergence descriptor, $\bm{b}$, by first creating an aggregate object class distribution. To create this aggregate distribution, we follow the work of~\cite{clemen1999combining}, insofar as each child capsule type is treated as an expert giving its prediction about the true underlying class distribution. First, we compute a simple linear opinion pool~\cite{stone1961opinion}, $\bm{p(\tilde z)} = \sum_i^N \sigma_s(\bm{z}_i)/N$, where $\bm{p(\tilde z)}$ is the aggregated probability distribution, $\bm{Z} = \{\bm{z}_1, \bm{z}_2, ..., \bm{z}_N\}$ is the set of child class presence projection vectors, and $\sigma_s(\bm{z}_i)_j = e^{\bm{z}_{ij}} / \sum_k^K e^{\bm{z}_{ik}} \textrm{ for } j = \{1, ..., K\}, i = \{1, ..., N\}$ is the softmax function used to transform projection vectors into normalized probability distributions over the $K$ classes. Then, we measure the agreement between each child's predicted distributions, $\sigma_s(\bm{z}_i)$, and the aggregate distribution, $\bm{p(\tilde z)}$, as the KL divergence between them $b_i = \sum_k^K p(\tilde z_k) \log ( p(\tilde z_k) / \sigma_s(z_i)_k)$.

\medskip

Lastly, we take child capsules' predicted distributions, $\sigma_s(\bm{z}_i)$, and compute their variance to estimate the uncertainty each child has: $c_i = \sum_k^K (\sigma_s(z_i)_k - \sum_k^K \sigma_s(z_i)_k)^2$. Our three sets of descriptors are efficiently computed for all capsules simultaneously (i.e. for entire batch and across spatial locations) on GPU in parallel with matrix operations. They are then concatenated, $\bm{s} = \bm{a} \oplus \bm{b} \oplus \bm{c}$, and fed to the excitation layers of our routing mechanism (Fig. 2).

%\medskip
%%%%%%%%%%%%%%%%%%%%%%%%%%%%%%%%%%%%%%%%%%%%%%%%%%%%%%%
\subsubsection{Determining routing coefficients (excitation):} 
%%%%%%%%%%%%%%%%%%%%%%%%%%%%%%%%%%%%%%%%%%%%%%%%%%%%%%% Don't add space below this line here!
The excitation stage of the SE-Routing algorithm has the task of learning a mapping from the concatenated set of capsule descriptors, $\bm{s}$, into a set of routing coefficients for the child capsule projections. Since parents capsules types are no longer different classes in our formulation, but rather two separated aspects of modeling objects, we compute a single set of routing coefficients at each spatial location for both parents. Formally, this mapping is computed as $\bm{r} = \sigma(\bm{W}_2\delta(\bm{W}_1\bm{s}))$, where $\bm{W}_1 \in \mathbb{R}^{\frac{3N}{t} \times 3N}$, $\bm{W}_2 \in \mathbb{R}^{N \times \frac{3N}{t}}$, $\delta$ is the ReLU activation function, $\sigma$ is the sigmoid activation function, and $t$ is the reduction ratio used to form this mapping into a two fully-connected (FC) layer bottleneck. A brief note: although excitation routing has interesting parallels to self-attention (e.g. dynamically conditioned on the input), our learned mapping is non-mutually-exclusive, while self-attention and CapsNet's dynamic routing both rely on applying a softmax function over outputs.

\medskip

Finally, with the determined routing coefficients $\bm{r} = \{r_1, r_2, ..., r_N\}$, we can compute the output of the SplitCaps detection head. Projection vectors from each child to each parent are computed using the proposed deformable capsules. These projections are then combined and weighted by the routing coefficients to form the final parent capsules. These final parents contain the instantiation parameters, $\bm{v}_{obj} = \sum_i^N r_i\bm{\hat u}_{obj|i}$, and class presence, $\bm{v}_{cls} = \sum_i^N r_i\bm{\hat u}_{cls|i}$, of any objects being represented at the given spatial location within the detection grid.

\medskip
\textbf{Top-level algorithm overview of DeformCaps:}  
Traditional capsule networks are limited in representing object deformations due to fixed spatial constraints. To solve this, we introduce Deformable capsules in this work. These capsules learn an additional set of weights that act as offsets for spatial sampling of child capsules. This allows them to adaptively gather information from non-fixed locations, effectively modeling object poses and deformations. Deformable capsules improve the ability to represent object variations while maintaining efficiency. While capsules are promising, scaling capsule networks to large datasets (like MS COCO) is challenging due to the high memory requirements for representing each class with a dedicated capsule type. To address this issue, herein we introduce SplitCaps, a new capsule structure, consisting of two types of parent capsules:
(1) Class-agnostic object instantiation capsules and class presence capsules. Class-agnostic capsules model the variations (pose, deformation, etc.) of objects, independent of class while class-presence capsules predict the presence of different object classes at a given location. (2) SE-Routing is a novel routing algorithm used to weight information coming from child capsules based on their agreement for each projected parent capsule. SplitCaps reduces memory consumption by separating class information from object instantiation parameters. SE-Routing ensures efficient information flow within the network. Overall DeformCaps architecture is based on CenterNet, which represents objects as point values at their bounding box centers. Features extracted from a CNN backbone (DLA-34) are fed into the capsule object detection head and a bounding box regression head. Then, deformable capsules are used to model object variations. Finally, SplitCaps and SE-Routing are employed for efficient object detection and class prediction.

%%%%%%%%%%%%%%%%%%%%%%%%%%%%%%%%%%%%%%%%
\section{Deformable capsules on MS COCO} \label{sec:exp}
%%%%%%%%%%%%%%%%%%%%%%%%%%%%%%%%%%%%%%%%
We evaluated our deformable capsule object detection framework on the MS COCO dataset~\cite{lin2014microsoft}, which contains $118$K training, $5$K validation and $20$K hold-out testing images (this data set is licensed under a Creative Common Attribution 4.0 International (CC BY 4.0)). When evaluating object detection models, we relied on metrics that go beyond simply counting how many objects are correctly identified. For this, we used Average Precision (AP) metric as a common metric that considers both precision (percentage of detection that are truly  the object of interest) and recall (percentage of actual objects that are detected). AP calculates the average precision across all possible detection confidence thresholds. A higher AP signifies a model that's good at finding most objects while minimizing false positives. We also used variations of APs. For instance, AP50 measures the average precision when the model only keeps detection with a confidence score of 50\% or higher. This helps assess the model's performance at stricter confidence levels. AP is reported over all IOU thresholds and at thresholds $0.5$ (AP50) and $0.75$ (AP75). We followed the training procedure as described in~\cite{zhou2019objects}, training on $512 \times 512$ pixel inputs, yielding $128 \times 128$ detection grids, using random flip, random scaling (between 0.6 to
1.3), cropping, and color jittering as data augmentation, and Adam~\cite{kingma2014adam} to optimize our objective function. Due to limited compute resources, we initialized the backbone network weights from CenterNet and train for $40$ epochs only with a batch size of $12$ and learning rate of 5e-4 with $5\times$ drops at $5$, $15$, and $25$ epochs. Longer training would likely yield superior results, as found by~\cite{zhou2019objects} who obtained better results for CenterNet when increasing from $140$ to $230$ epochs. 

In Table~\ref{table:results}, we provide results of our proposed deformable capsule network with and without flip and multi-scale augmentations following~\cite{zhou2019objects}. Inference time on our hardware (Intel Xeon E5-2687 CPU, Titan V GPU, Pytorch 1.2.0, CUDA 10.0, and CUDNN 7.6.5) was consistent with those reported by ~\cite{zhou2019objects}.\footnote{Our code is publicly available for the community for reproducible research.} In single-stage detector methods (middle row), \textit{DeformCaps} outperforms all the other available methods in the literature, and on par results with CenterNet. Specifically, \textit{DeformCaps} performs slightly inferior to the CenterNet in terms of AP, but produces far fewer false positive detection, as shown in Table~\ref{table:results}. Note that one-stage detectors offer advantages like speed and simplicity, but they can sometimes fall short in accuracy compared to two-stage detectors. It is important to note that \textit{DeformCaps} is a single-stage detector, and not expected to perform better than two-stage detectors as it is kept simple and efficient. It is possible to convert a one-stage detector into a two-stage detector if accuracy is of a significant concern. This can be done by leveraging dedicated region proposal generation, more focused feature extraction, and a reduced classification bruden on the network. However, this comes at the cost of increased complexity and potentially slower processing speeds.

\medskip
For ablations (last two rows in Table 1), we trained a version of \textit{DeformCaps} which replaces the proposed deformable capsules with the standard locally-constrained convolutional capsules (\textit{non-DeformCaps}), and a version which removed the routing procedures (\textit{No-Routing}).
%, and one which removed the reconstruction regularization sub-network (\textit{No-ReconReg}).
These ablations show the contribution of each component of the proposed method. Proposed method with both routing mechanism and deformable capsules provided better accuracy as shown in Table 1.

\begin{table*}[h]
\caption{Single-scale / multi-scale results on COCO test-dev. Top: Two-stage detectors; Middle: One-stage detectors; Bottom: Ablation studies on deformable capsules and SE-Routing. Two lines (\textit{DeformCaps} and CenterNet) are highlighted.\label{table:results}}
{\resizebox{\columnwidth}{!}{
\begin{tabular}{|l|l|c|c|c|c|c|c|c|}
\hline
\textbf{Method} & \textbf{Backbone} & \textbf{FPS} & \textbf{AP} & \textbf{AP50} & \textbf{AP75} & \textbf{APS} & \textbf{APM} & \textbf{APL} \\ 
\hline
MaskRCNN~\cite{he2017mask}    & ResNeXt-101       & 11  & 39.8 & 62.3 & 43.4 & 22.1 & 43.2 & 51.2  \\
Deform-v2~\cite{zhu2019deformable}   & ResNet-101        & -   & 46.0 & 67.9 & 50.8 & 27.8 & 49.1 & 59.5  \\
SNIPER~\cite{singh2018sniper}      & DPN-98            & 2.5 & 46.1 & 67.0 & 51.6 & 29.6 & 48.9 & 58.1  \\
PANet~\cite{liu2018path}       & ResNeXt-101       & -   & 47.4 & 67.2 & 51.8 & 30.1 & 51.7 & 60.0  \\
TridentNet~\cite{li2019scale}  & ResNet-DCN    & 0.7 & 48.4 & 69.7 & 53.5 & 31.8 & 51.3 & 60.3  \\ 
\hline
YOLOv3 (Redmon et al., 2018)     & DarkNet-53        & 20  & 33.0 & 57.9 & 34.4 & 18.3 & 25.4 & 41.9  \\
RetinaNet~\cite{lin2017focal}   & ResNeXt-FPN   & 5.4 & 40.8 & 61.1 & 44.1 & 24.1 & 44.2 & 51.2  \\
RefineDet~\cite{zhang2018single}   & ResNet-101        & -   & 36.4 / 41.8 & 57.5 / 62.9 & 39.5 / 45.7 & 16.6 / 25.6 & 39.9 / 45.1 & 51.4 / 54.1   \\
CenterNet~\cite{zhou2019objects}   & \textbf{DLA-34}            &\textbf{ 28}  & \textbf{39.2} / \textbf{41.6} &\textbf{ 57.1} / \textbf{60.3} & \textbf{42.8} / \textbf{45.1} & \textbf{19.9} / \textbf{21.5} & \textbf{43.0} / \textbf{43.9} & \textbf{51.4} / \textbf{56.0}      \\
\textbf{DeformCaps (Ours)} & \textbf{DLA-34} & \textbf{15} & \textbf{38.0} / \textbf{40.6} & \textbf{55.5} / \textbf{58.6} & \textbf{41.3} / \textbf{43.9} & \textbf{20.0} / \textbf{23.0} & \textbf{42.5} / \textbf{42.9} & \textbf{51.8} /\textbf{ 56.4} \\
\hline
Non-DeformCaps      & DLA-34    & 20 & 27.2 / 30.1 & 41.8 / 45.9 & 29.6 / 32.6 & 13.3 / 15.6 & 30.0 / 30.8 & 35.0 / 40.6 \\ 
No-Routing          & DLA-34    & 15 & 35.9 / 38.9 & 52.7 / 56.0 & 39.1 / 42.4 & 17.8 / 20.9 & 40.6 / 42.0 & 48.2 / 54.4 \\ \hline
\toprule
\end{tabular}
}}
\end{table*}

\subsection*{Qualitative Results on MS COCO} 
\begin{figure*}
\centering
\includegraphics[width=\textwidth]{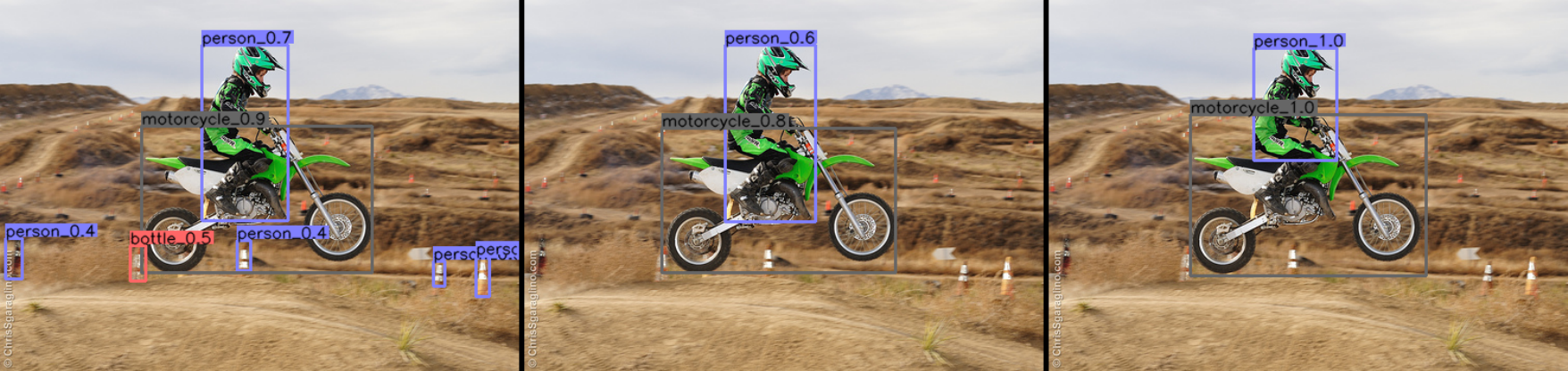}
\caption{Qualitative example for CenterNet~\cite{zhou2019objects} (\textbf{leftmost}), the proposed DeformCaps (\textbf{center}), and the ground-truth annotations (\textbf{rightmost}) on the MS COCO test-dev dataset. While CenterNet produces slightly higher average precision (AP) values than DeformCaps, it also seems to produce more false positives on average than DeformCaps.} \label{fig:qual_single}
\end{figure*}

Included in this section are a number of qualitative examples for CenterNet~\cite{zhou2019objects} and the proposed DeformCaps on the MS COCO test-dev dataset~\cite{lin2014microsoft} using both flip and multi-scale augmentations. Results for CenterNet were obtained using the official code and trained models provided by the authors~\cite{zhou2019objects}. While we do not make any sweeping claims, we wish to comment on a few general patterns that seemed to emerge in these examples. In Figure~\ref{fig:qual_single}, we show a prototypical example of the general trend we are describing. DeformCaps' results (middle) are more close to ground truth (right), and superior to CenterNet's results (left) in terms of less false positive, AP is higher in CenterNet results. AP precisions are close to each other, though.

\begin{figure*}
  \centering
  \includegraphics[width=0.9\linewidth]{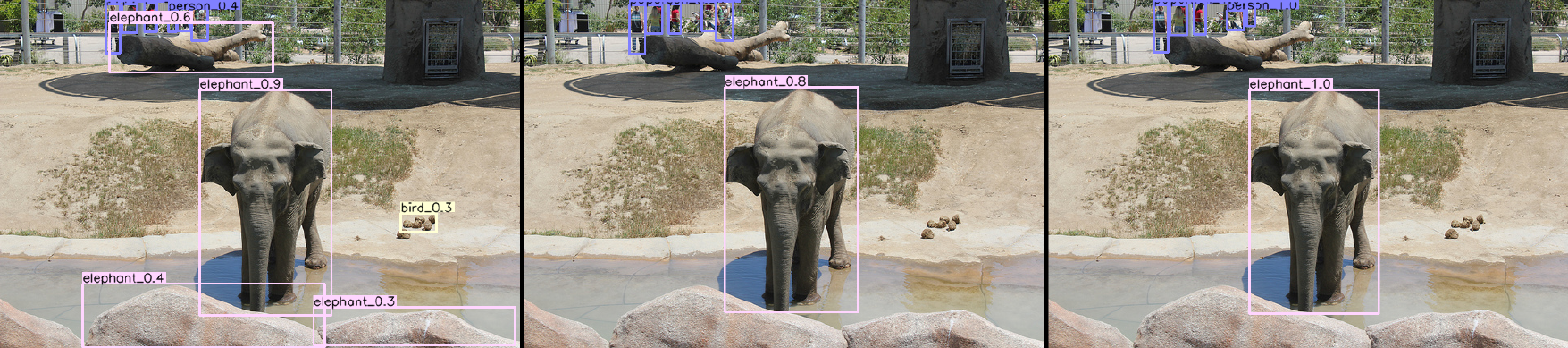}
  \label{fig:fp_sub_1}
  \centering
  \includegraphics[width=0.9\linewidth]{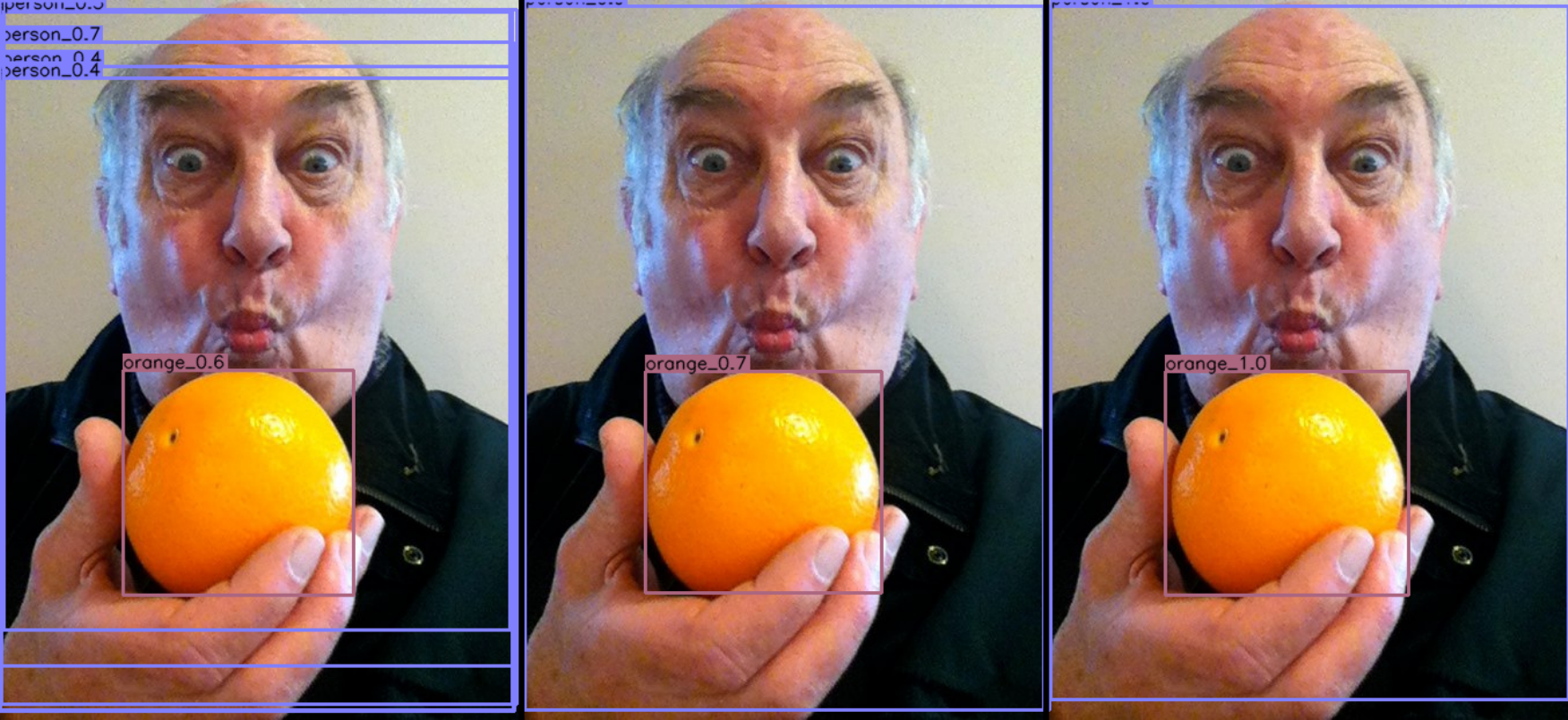}
  \label{fig:fp_sub_2}
    \centering
  \includegraphics[width=0.9\linewidth]{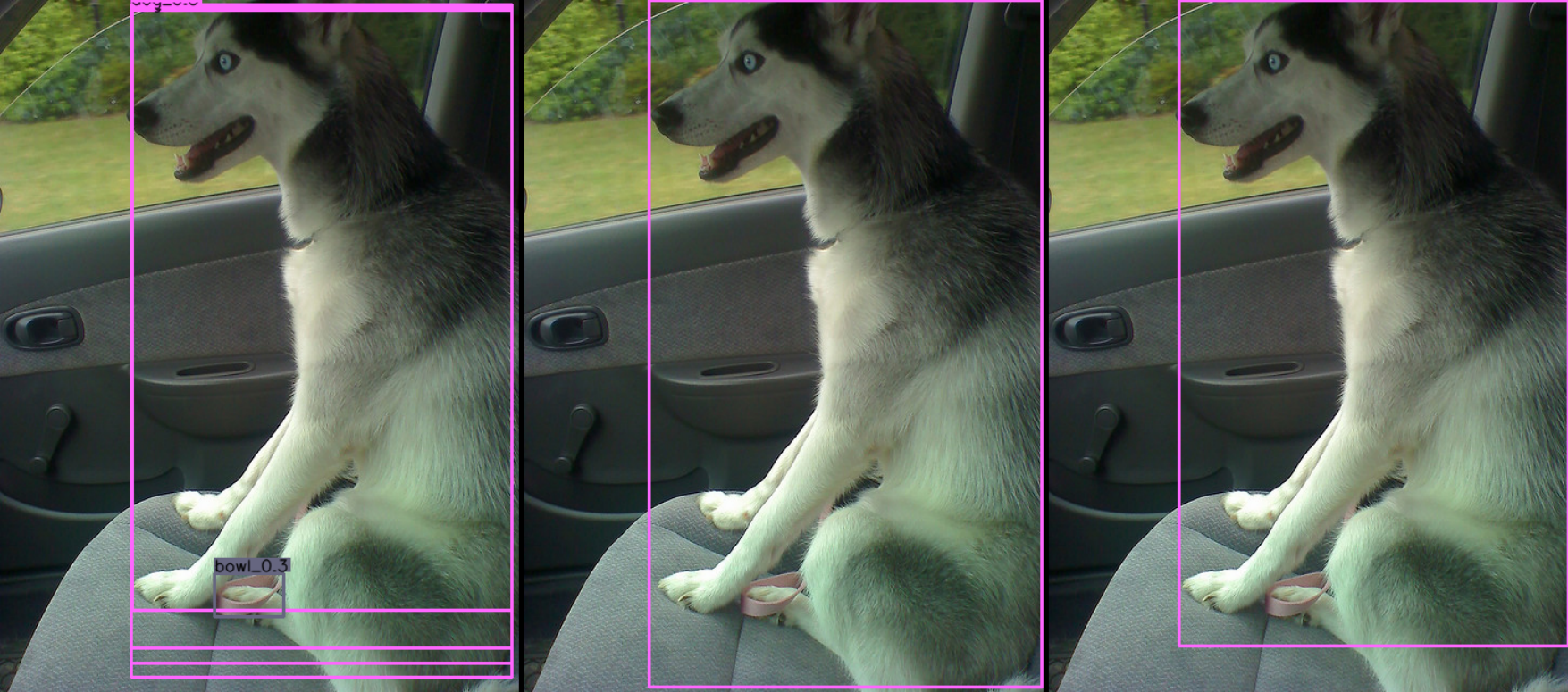}
  \label{fig:fp_sub_3}
  \includegraphics[width=0.9\linewidth]{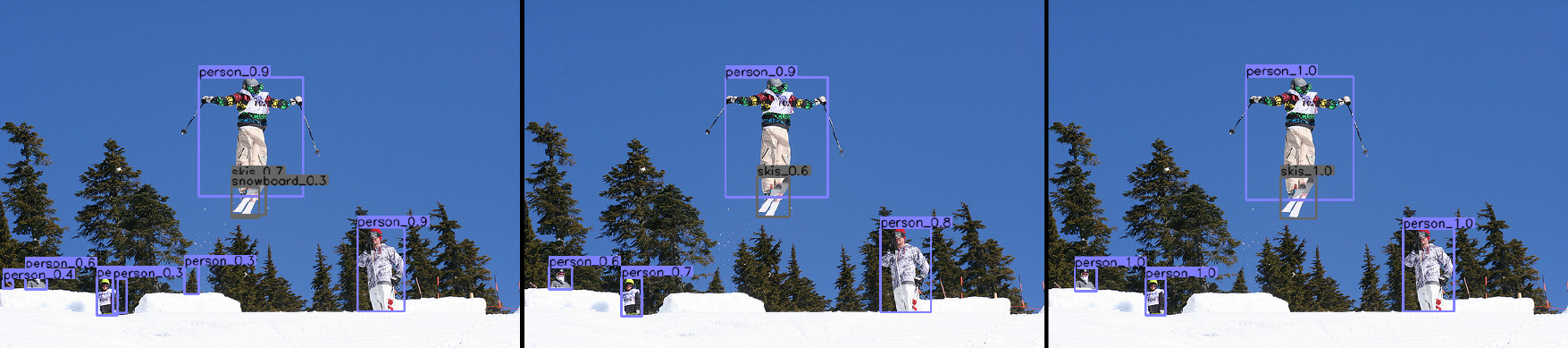}
  \label{fig:fp_sub_4}
\caption{CenterNet~\cite{zhou2019objects} (\textbf{leftmost}) consistently produces a higher number of \textit{false positives} than the proposed \textit{DeformCaps} (\textbf{center}), as compared to the ground-truth annotations (\textbf{rightmost}) on the MS COCO test-dev dataset.}
\label{fig:qual_fp}
\end{figure*}

\begin{figure*}
  \includegraphics[width=0.9\linewidth]{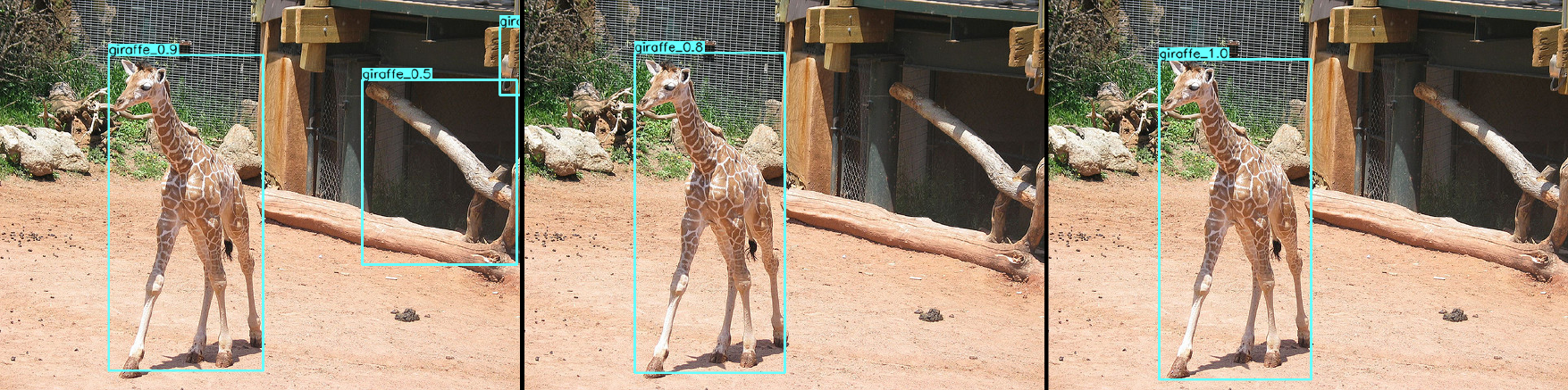}
  \label{fig:fp2_sub_1}

  \includegraphics[width=0.9\linewidth]{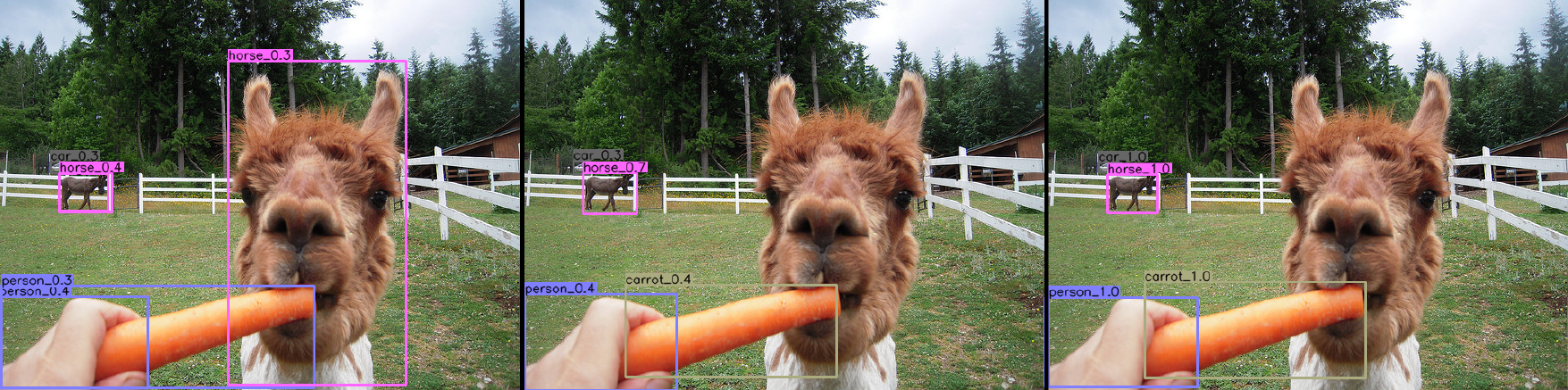}
  \label{fig:fp2_sub_2}

\includegraphics[width=0.9\linewidth]{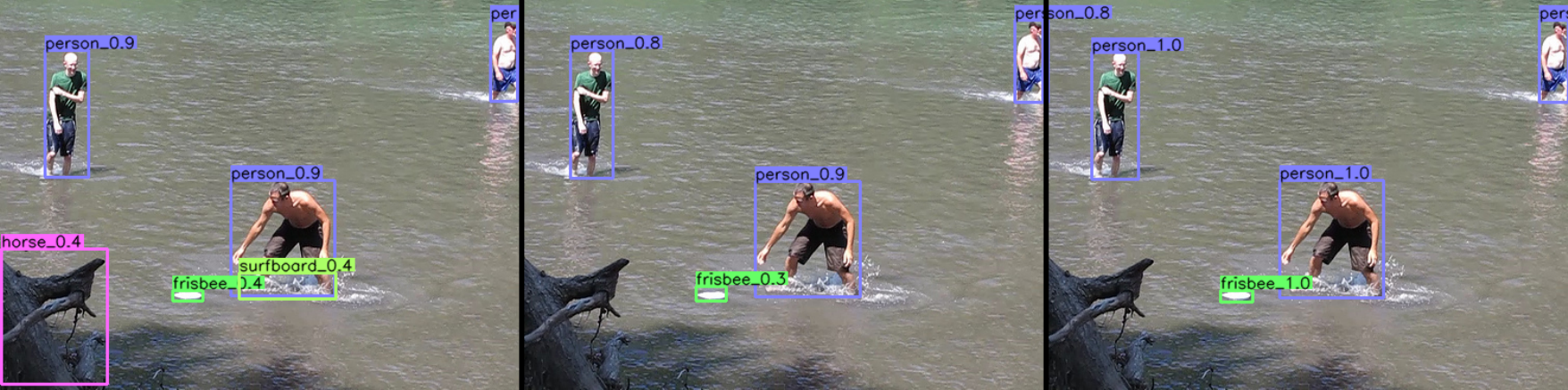}
  \label{fig:fp2_sub_3}
  \includegraphics[width=0.9\linewidth]{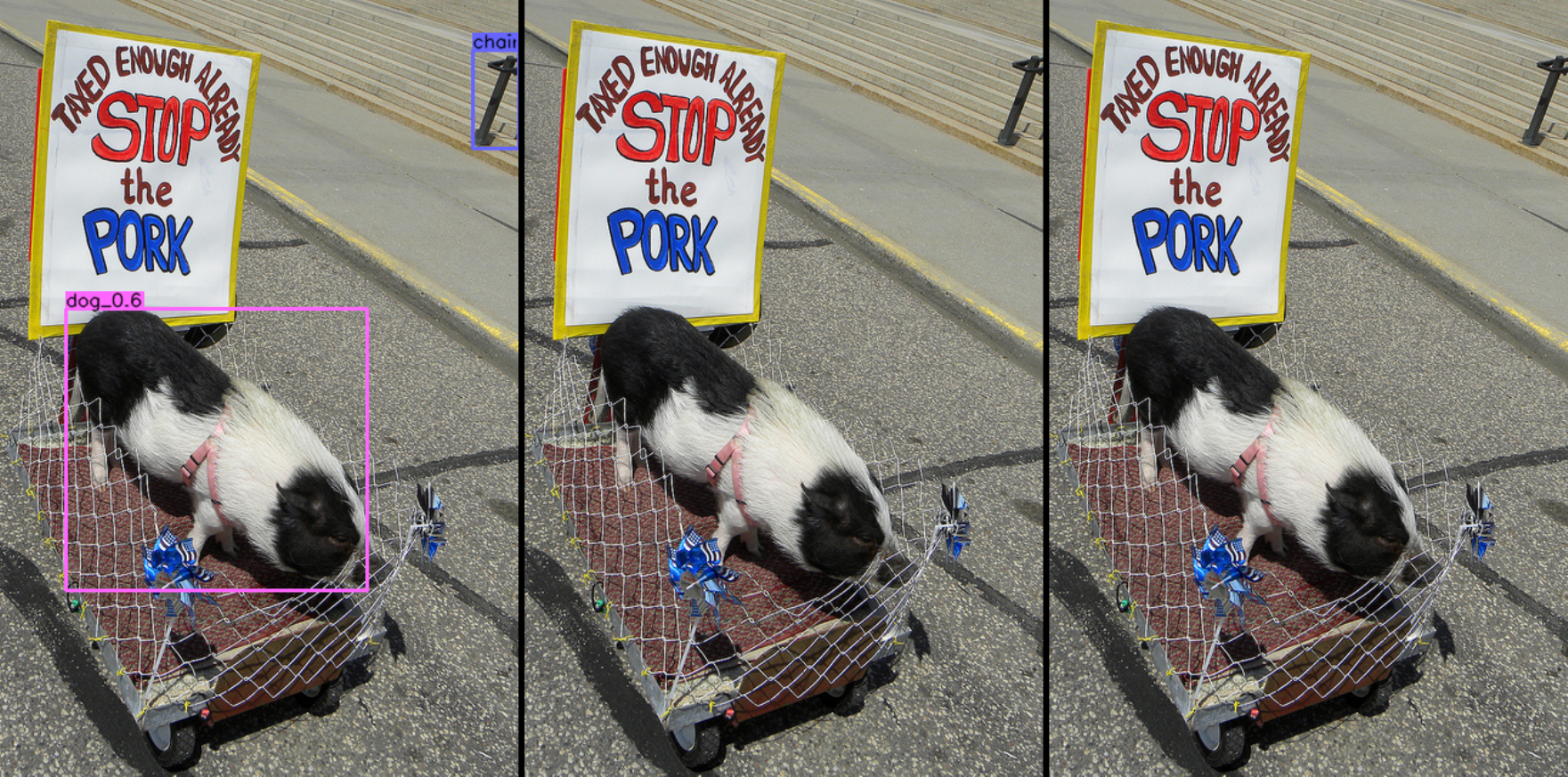}
  \label{fig:fp2_sub_4}
\caption{Some more qualitative examples showing CenterNet~\cite{zhou2019objects} (\textbf{leftmost}) consistently producing a higher number of \textit{false positive detections} than the proposed DeformCaps (\textbf{center}), as compared to the ground-truth annotations (\textbf{rightmost}) on the MS COCO test-dev dataset.}
\label{fig:qual_fp2}
\end{figure*}

\begin{figure*}
  \centering
  \includegraphics[width=\linewidth]{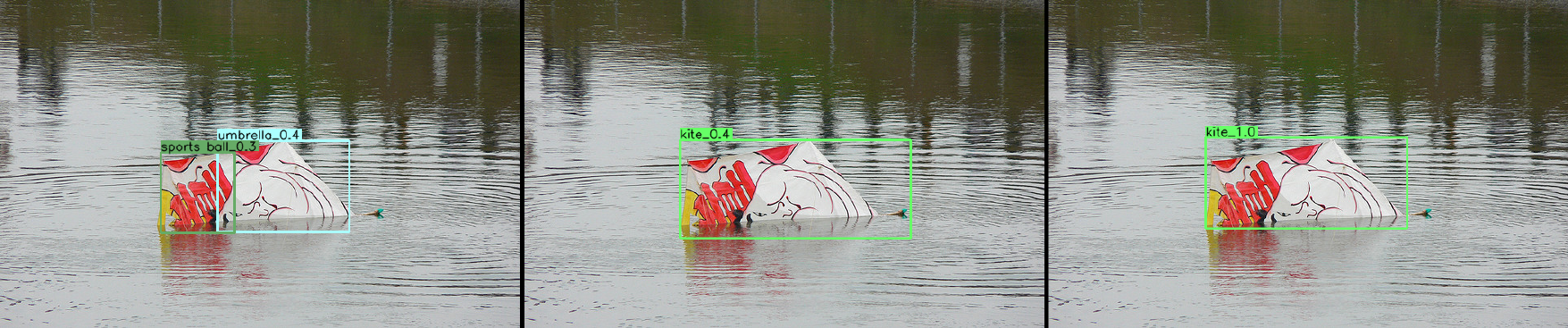}
  \label{fig:fn2_sub_1}

  \centering
  \includegraphics[width=\linewidth]{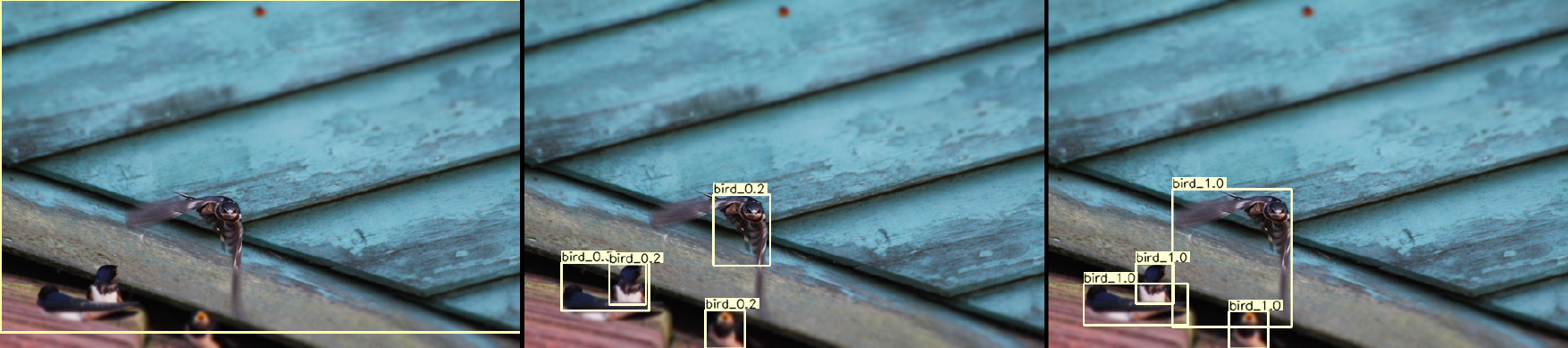} 
  \label{fig:fn2_sub_2}

  \centering
  \includegraphics[width=\linewidth]{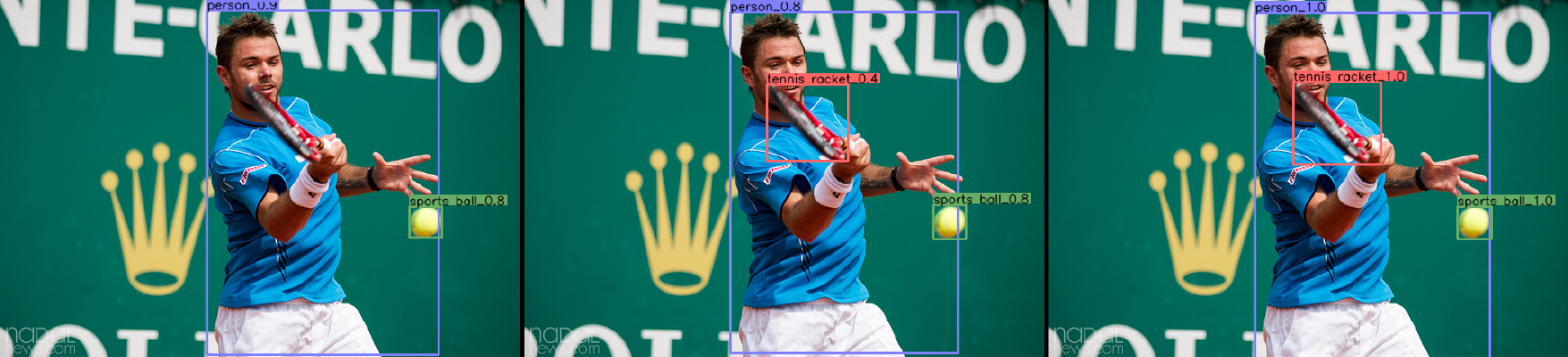} 
  \label{fig:fn2_sub_3}

  \centering
  \includegraphics[width=\linewidth]{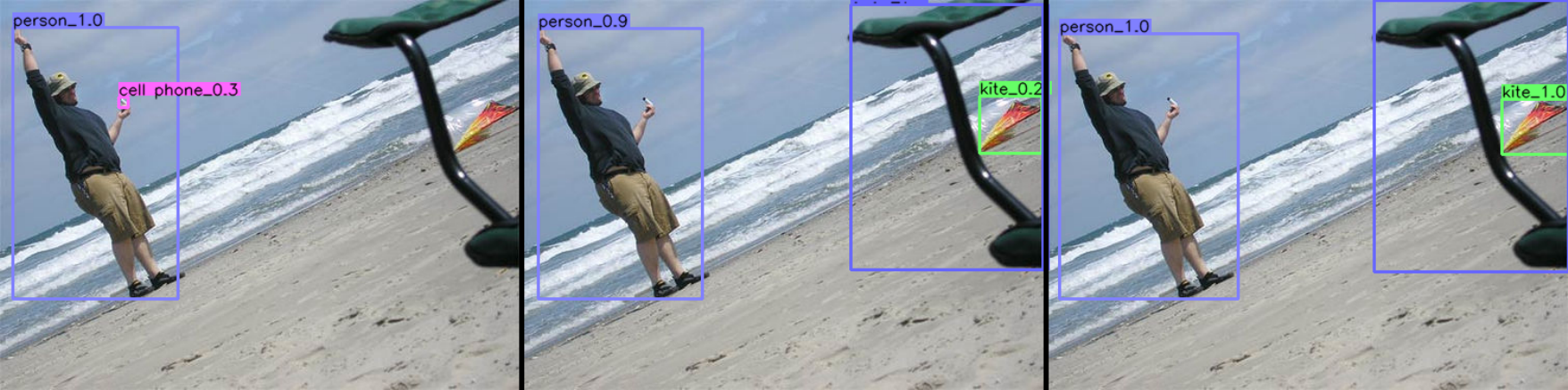} 
  \label{fig:fn2_sub_4}

  \centering
  \includegraphics[width=\linewidth]{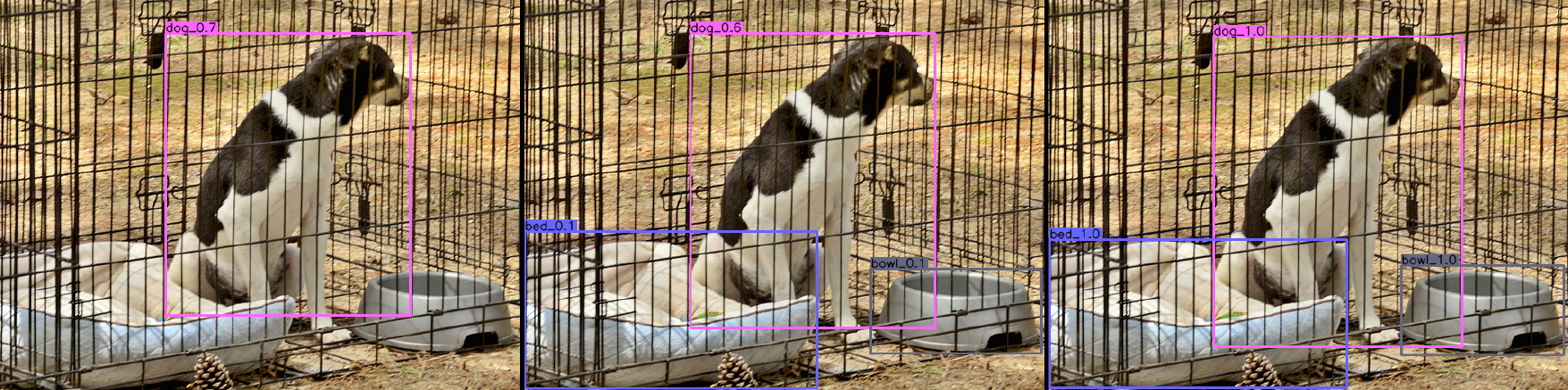}
  \label{fig:fn2_sub_5}

\caption{Qualitative examples showing CenterNet~\cite{zhou2019objects} (\textbf{leftmost}) struggling to detect objects being presented \textit{in unusual poses or from unusual viewpoints}, while the proposed \textit{DeformCaps} (\textbf{center}) successfully captures these cases, as compared to the ground-truth annotations (\textbf{rightmost}) on the MS COCO test-dev dataset.}
\label{fig:qual_fn}
\end{figure*}

\begin{figure*}
  \centering
  \includegraphics[width=\linewidth]{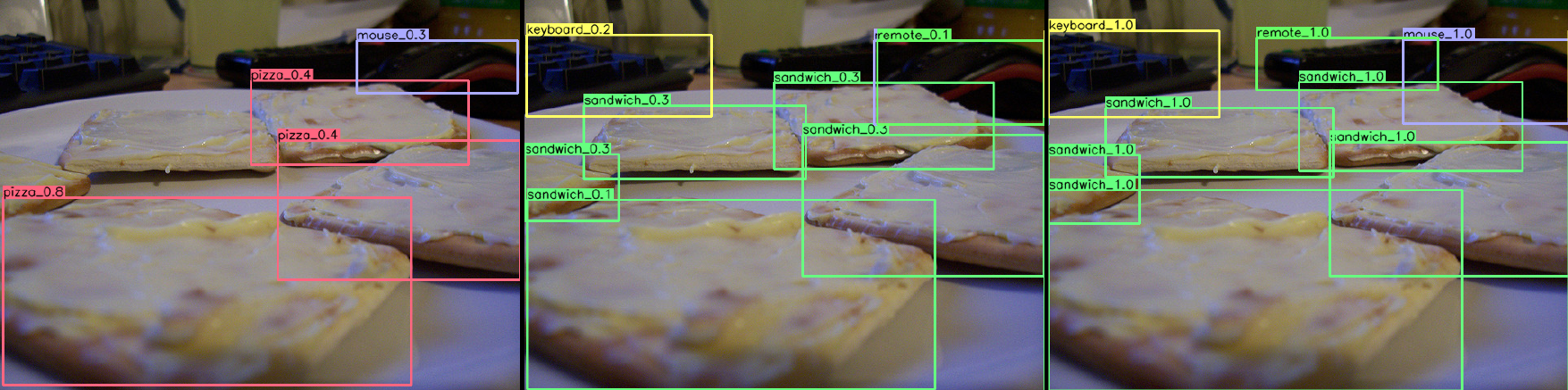}
  \label{fig:fn_sub_1}

  \centering
  \includegraphics[width=\linewidth]{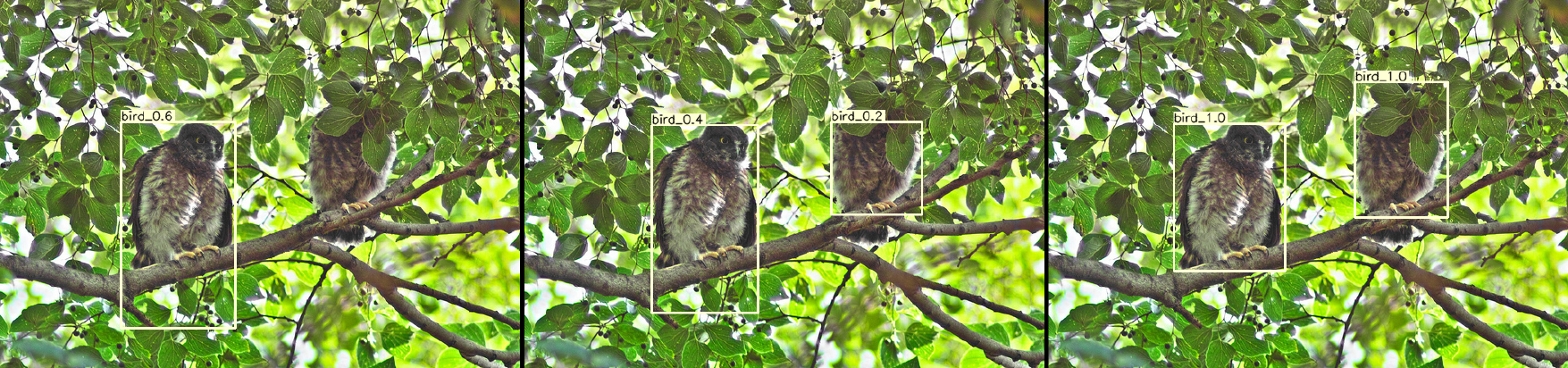}
  \label{fig:fn_sub_2}

  \centering
  \includegraphics[width=\linewidth]{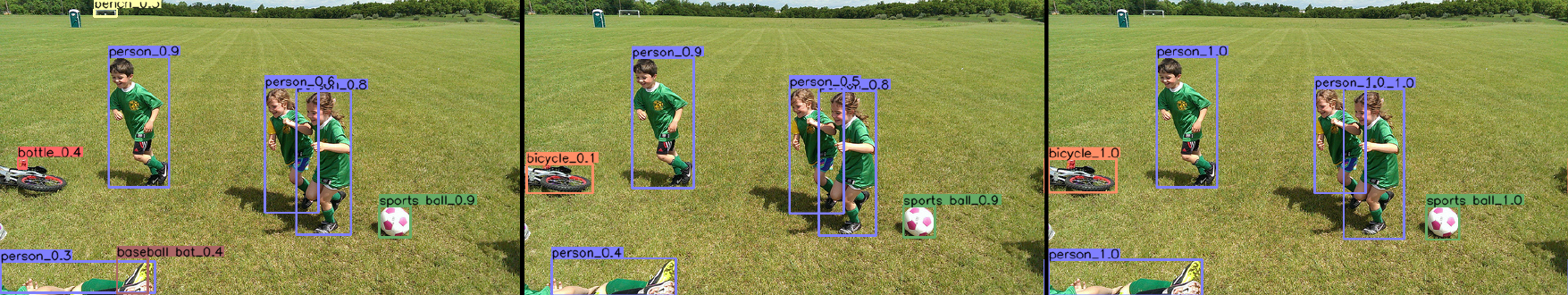} 
  \label{fig:fn_sub_3}

  \centering
  \includegraphics[width=\linewidth]{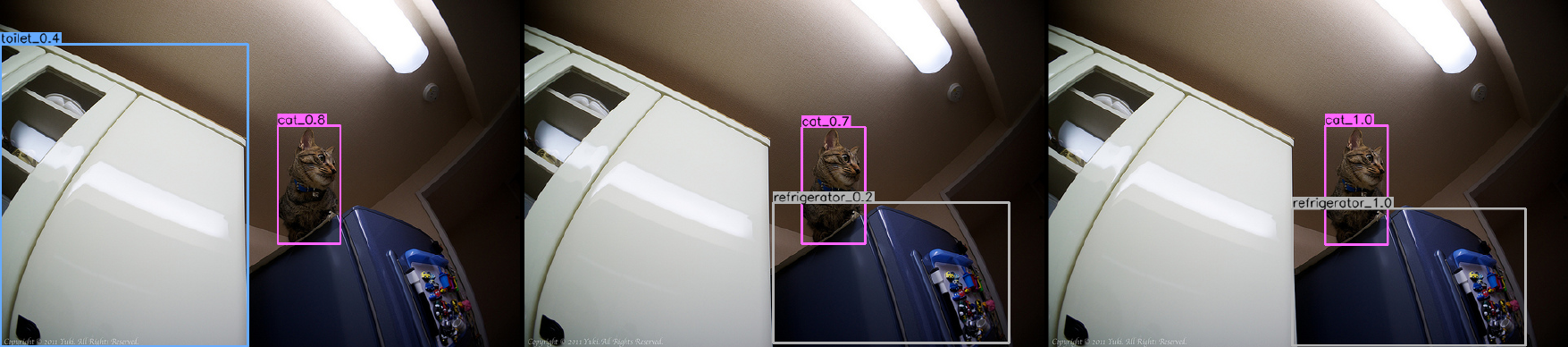}
  \label{fig:fn_sub_4}

  \centering
  \includegraphics[width=\linewidth]{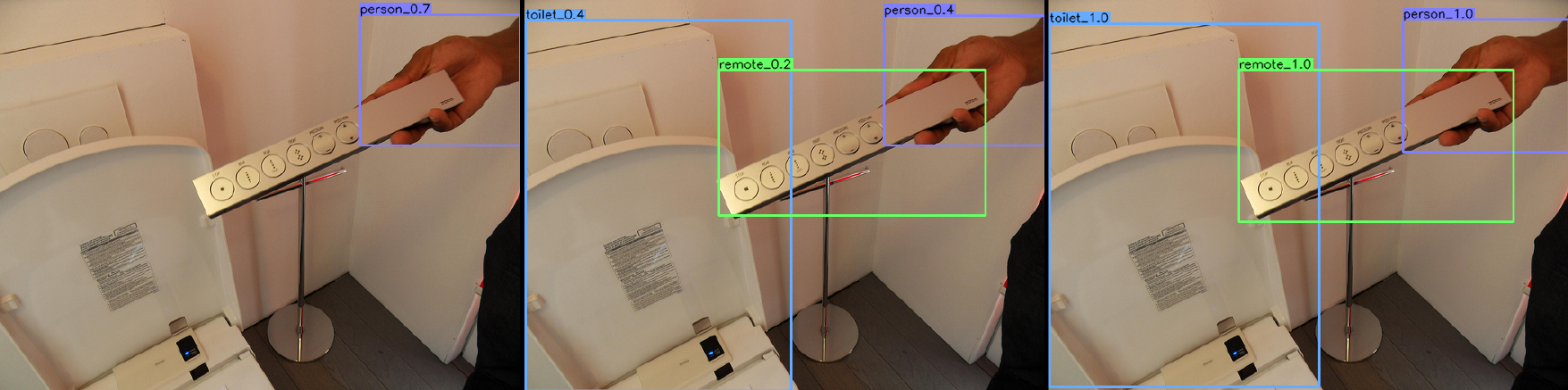}
  \label{fig:fn_sub_5}

\caption{A few more qualitative examples showing CenterNet~\cite{zhou2019objects} (\textbf{leftmost}) struggling to detect objects being presented in \textit{in unusual poses or from unusual viewpoints}, while the proposed \textit{DeformCaps} (\textbf{center}) successfully captures these cases, as compared to the ground-truth annotations (\textbf{rightmost}) on the MS COCO test-dev dataset.}
\label{fig:qual_fn2}
\end{figure*}

In Figures~\ref{fig:qual_fp}--~\ref{fig:qual_fp2}, we include more examples of this trend. In Figure 4, for instance, CenterNet method consistently produces a higher number of false positives than the proposed DeformCaps (center), as compared to the ground-truth annotations (rightmost) on the MS COCO test-dev dataset. In Figures~\ref{fig:qual_fn}--~\ref{fig:qual_fn2}, we include a set of interesting examples of unusual object viewpoints or poses being better captured by DeformCaps than by CenterNet. Figure 6's last row has some results where CenterNet has a missing object, while DeformCaps catch the object of interests (bed and bowl) unlike other examples in Figures 4, 5, and 6. In Figure 7, we observed similar strugles of CenterNet when capturing objects with unusual poses or viewpoints while the proposed DeformCaps (center) successfully captures these cases,
as compared to the ground-truth annotations (rightmost) on the MS COCO test-dev dataset.

First, DeformCaps tends to be more conservative with its detections than CenterNet. This can be observed both by the slightly lower confidence scores (typically 0.1 less than CenterNet for most detections), and by the overall fewer amount of boxes placed in scenes. Second, CenterNet tends to produce far more false positives than DeformCaps, both in the case of incorrect detections and of multiple detections for the same object which failed to be suppressed by the NMS algorithm. Though DeformCaps producing slightly lower confidence scores might account for some of the reduction in false positives, we observe CenterNet consistently producing fairly confident false predictions (e.g. $> 0.4$) while DeformCaps does not produce a detection in the same region at all.

This observation of less false positive detection is consistent with what we would expect from a capsule network with dynamic routing as compared to a traditional convolutional neural network (CNN). Where a CNN passes on all activations to the next layer, capsule networks utilize a dynamic routing algorithm to only pass on activations if there is agreement amongst the child capsule projections. In our proposed method specifically, with a SplitCaps structure and SE-Routing, the agreement is computed for projections of both the pose and class of the object being represented. It follows naturally that this would limit the amount of false positive detection which are produced, by reducing the amount of activations that get passed on. Further, we find from a survey of these qualitative examples that DeformCaps is better able to detect objects when being presented in an unusual pose or from an usual viewpoint than its CenterNet counterpart. This gives empirical support to one of the purported benefits of capsule networks, to be able to better generalize to unseen poses and viewpoints.

%We further include a number of qualitative examples for CenterNet~\cite{zhou2019objects} and the proposed DeformCaps on the MS COCO test-dev dataset~\cite{lin2014microsoft} using both flip and multi-scale augmentations. Results for CenterNet were obtained using the official code and trained models provided by the authors~\cite{zhou2019objects}. While we do not make any sweeping claims, we wish to comment on a few general patterns that seemed to emerge in these examples. In Figure~\ref{fig:qual_single}, we show a prototypical example of the general trend we are describing. In Figures~\ref{fig:qual_fp}--~\ref{fig:qual_fp2}, we include more examples of this trend. In Figures~\ref{fig:qual_fn}--~\ref{fig:qual_fn2}, we include a set of interesting examples of unusual object viewpoints or poses being better captured by DeformCaps than by CenterNet.

%%%%%%%%%%%%%%%%%%%%%%%
\section{Related works} \label{sec:related}
%%%%%%%%%%%%%%%%%%%%%%%

%%%%%%%%%%%%%%%%%%%%%%%%%%%%%
\subsection{Capsule networks}  \label{sec:related_caps}
%%%%%%%%%%%%%%%%%%%%%%%%%%%%%
Capsule neural networks have been proposed as a stronger alternative to CNNs for learning object-centric representations, which play a critical role in better generalization and sample complexity. The idea of capsules was first introduced by~\cite{hinton2011transforming}. \cite{sabour2017dynamic} extended this and proposed dynamic routing between capsules. The EM routing algorithm was then modified by~\cite{hinton2018matrix}. Capsules are designed explicitly model part-whole relationship by using a group of neurons to encode visual entities and learn the relationship between these entities~\cite{survey}. Recently, capsule networks could achieve the state-of-the-art performance for a wide range of applications: video object segmentation~\cite{duarte2019capsulevos}, point cloud segmentation~\cite{zhao20193d}, explainable medical diagnosis~\cite{lalonde2020encoding}, text classification~\cite{zhao2018investigating}, sentiment analysis~\cite{wang2018sentiment}, and various other applications~\cite{vijayakumar2019comparative}. Unlike high precision and accuracy that recent studies obtained in various applications, capsule networks do not yet work efficiently, which can in part be attributed to their lack of efficiency. The complexity induced by vector valued neural activities and capsule routing algorithm lead to very inefficient models that are often difficult to train. The current study in this paper partially addresses this difficulty for the first time in the literature. For a comprehensive review, a recent survey paper~\cite{survey} throughly provides detailed breakdown of capsule networks and related research on object-centric representation learning.

%%%%%%%%%%%%%%%%%%%%%%%%%%%%%
\subsection{Object detection}  \label{sec:related_det}
%%%%%%%%%%%%%%%%%%%%%%%%%%%%%

%%%%%%%%%%%%%%%%%%%%%%%%%%%%%%%%%%%%%%%%%%
\textbf{Region proposal-based approaches:}
%%%%%%%%%%%%%%%%%%%%%%%%%%%%%%%%%%%%%%%%%% Don't add space below this line here!
R-CNN was one of the first successful deep object detectors, in which a selective search algorithm was used to select a number of region proposals, CNN features were extracted from each of the region proposals and were used to both classify the object and regress its bounding box~\cite{girshick2014rich}. The later addition of Fast R-CNN~\cite{girshick2015fast} provided end-to-end training and addressed the speed and efficiency issues of R-CNN.

%%%%%%%%%%%%%%%%%%%%%%%%%%%%%%%%%%
\textbf{Anchors-based approaches:}
%%%%%%%%%%%%%%%%%%%%%%%%%%%%%%%%%% Don't add space below this line here!
Anchors-based approaches sample fixed-shape bounding boxes (anchors) around a low-resolution image grid, then attempt to classify anchors into object classes. Faster R-CNN~\cite{ren2015faster} generates region proposals in a first stage network, then attempts to classify and regress bounding boxes for the top-$k$ highest scoring anchors in a second stage network. Later studies such as~\cite{redmon2016you} dramatically speed up the process by converting the proposal classifier to a multi-class one-stage detector. Since then, researchers have been working on improving one-stage detectors by including shape priors~\cite{redmon2017yolo9000,redmon2018yolov3}, multiple feature resolutions~\cite{liu2016ssd}, re-weighting the loss among different samples~\cite{lin2017focal}, or modeling channel-wise attention~\cite{chen2020recursive}.

%%%%%%%%%%%%%%%%%%%%%%%%%%%%%%%%%%%%%%%%%%%%%%
\textbf{Keypoint estimation-based approaches:}
%%%%%%%%%%%%%%%%%%%%%%%%%%%%%%%%%%%%%%%%%%%%%% Don't add space below this line here!
CornerNet~\cite{law2018cornernet} attempts to detect objects by predicting two bounding box corners as keypoints. ExtremeNet~\cite{zhou2019bottom} extends CornerNet's approach by estimating all corners and the center of the objects' bounding box. However, these methods rely on significantly slow combinatorial grouping post-processing stage. \cite{zhou2019objects} proposed CenterNet which attempts to predict only an objects' center point, and regress all other necessary values from there without the need for grouping or post-processing.

%%%%%%%%%%%%%%%%%%%%%%%%%%%%%
\subsection{How Deformable Capsules Address the Drawbacks of Conventional Methods?}  \label{sec:related_det}
Our proposed architecture represents a significant advancement in object detection, implementing a one-stage detection framework that achieves results comparable to state-of-the-art one-stage CNN-based methods on the MS-COCO dataset. The key innovations of our approach lie in its ability to reduce false positive detections while simultaneously improving generalization to objects in unusual poses or viewpoints.

One-stage detection frameworks, as opposed to two-stage methods, perform object localization and classification in a single forward pass through the network. This approach typically offers faster inference times, making it more suitable for real-time applications. Our method maintains this speed advantage while addressing some of the common shortcomings of one-stage detectors. Our proposed architecture also demonstrates the following other improvements over conventional methods.

\textbf{Reduced False Positives:} One of the primary challenges in object detection is achieving a balance between sensitivity (detecting all relevant objects) and specificity (avoiding false detections). Our deformable capsule method significantly reduces false positive detections, which is crucial for practical applications where false alarms can be costly or disruptive.

\textbf{Improved Generalization:} Object detection systems often struggle with objects in unusual poses or viewpoints that deviate from the most common presentations in the training data. Our deformable capsule architecture shows enhanced ability to correctly identify and localize objects in these challenging scenarios, indicating improved robustness and generalization capabilities.

\textbf{Large Training Data Requirements:} Current detection models require large amounts of labeled data to achieve high performance. Collecting and annotating such datasets can be time-consuming and expensive. On the other hand, Capsule networks are designed to understand spatial hierarchies between simple and complex objects. This ability allows them to generalize better from fewer examples compared to traditional CNNs, potentially reducing the dependency on large datasets.

\textbf{Lack of Interpretability:} Conventional deep learning detectors often function as "black boxes," making it challenging to understand how they make decisions. This lack of interpretability can be problematic for debugging, trust, and regulatory compliance. One of the strengths of capsule networks is their ability to retain information about the pose and spatial relationships of objects. This makes the models inherently more interpretable, as they preserve more detailed information about the objects and their parts.

\textbf{Sensitivity to Input Variations:} Classical deep learning detection models can be sensitive to variations in the input data, such as changes in lighting, occlusions, or distortions. Capsule networks are, on the other hand, designed to be more robust to variations such as rotations and affine transformations. They can better generalize across different input conditions due to their ability to understand spatial hierarchies and relationships.

%%%%%%%%%%%%%%%%%%%%%%%%%%%%%

%%%%%%%%%%%%%%%%%%%%%%%%%%%%%%%%%%%%%%%%%%%%%%%%%%
\section{Discussions and Conclusions} \label{sec:disc}
%%%%%%%%%%%%%%%%%%%%%%%%%%%%%%%%%%%%%%%%%%%%%%%%%%

Our proposed  (\textit{DeformCaps}) with \textit{SplitCaps} object-class representations and Squeeze-and-Excitation inspired \textit{SE-Routing} algorithm represents an important step for capsule networks to scale-up to large-scale computer vision problems, such as object detection or large-scale classification. Our proposed one-stage object detection capsule network is able to obtain results on MS COCO which are on-par with other state-of-the-art one-stage CNN-based networks for the first time in the literature, while also producing fewer false positives. Examining the qualitative results lends empirical evidence that \textit{DeformCaps} can better generalize to unusual poses/viewpoints of objects than CenterNet~\cite{zhou2019objects}. We hope our work will inspire future research into the considerable potential of capsule networks. 

It is important to note that our work is the first-ever capsule network for object detection and large-scale vision problems in general, requiring three major contributions to achieve; it is typical of such works to under-perform state-of-the-art. Further, it might be worth to note that some of the main benefits of capsules (\textit{e.g.} robustness to viewpoint/pose) are not captured well by standard AP metrics, where objects in unusual poses represent a very small fraction of examples.  

\textbf{Limitations:} Our study has some limitations too. Briefly,\\ 
\textbf{(1)} we had difficulty in integrating the bounding box regression values into our capsule object detection head. In our implementation, the class-agnostic capsules are trained to predict scale-normalized masks of $28 \times 28$. Ultimately, we would like to integrate predicting the object masks and the boxes for those masks together, as these tasks surely share mutual information. However, to the best of our knowledge, no published works exist for using capsules on a real-valued regression task.

\textbf{(2)} For our proposed SE-Routing, as with the original Squeeze-and-Excitation network, the choice of descriptors computed in the squeeze is somewhat handcrafted. We propose to use the cosine angle, KL divergence, and variance, and provide justifications for each of these choices, then allow the excitation to learn which of these pieces of information is most beneficial dynamically for each given input. Nonetheless, it is completely plausible that different descriptors could yield superior results. We unfortunately do not have the compute resources to run ablation studies over each of these chosen descriptors individually.

\textbf{(3)} The choice of 64 dimensions to model the class-agnostic instantiation parameters was decided empirically. As we argued previously, it is unlikely that all variations across object poses are completely class independent; thus, to represent these extra dimensions of variation, we increase our vector lengths considerably ($16 \rightarrow 64$). However, it is possible that the number of class-independent and class-dependent variations is significantly higher or lower than the value chosen, and largely will depend on the complexity of the data being modeled. This difficulty is analogous to determining the optimal number of convolutional filters to use at every given layer of a CNN. Related to this, there is the potential for the class-dependent dimensions of the instantiation vectors to have unwanted influence over the cosine angle descriptors when attempting to represent objects of other classes. It could be beneficial to pass class information from the class presence capsule type over to the object instantiation capsule type to dynamically attend to the relevant dimensions of its vector for a given object. In a similar manner, it could be beneficial when computing the probability aggregation using the linear opinion pool to weight the expert opinions in proportion to their uncertainty instead of uniformly. 

\textbf{(4)} We chose to reconstruct object's masks with the motivation of forcing the network to learn variations in shape, pose, and deformations. Since CNNs are known to be biased to texture information over shape, we chose not to explicitly supervise the learning of any texture information. Nonetheless, it is plausible that reconstructing the object with texture could yield superior performance. Further, we chose to set the value of the reconstruction regularization's contribution to the loss to $0.1$, following what was found most beneficial by CenterNet~\cite{zhou2019objects} for weighting the size loss contribution, and from a concern to not over-regularize the network early in training, then stepped this value to $2.0$ half-way through training to make its value roughly equal to the other loss terms. From our experience, the accuracy remained fairly consistent across values up to $2.0$ for this term, while setting its weight to $0.0$ resulted in a degradation of performance. We found that increasing the value during training led to faster improvements in performance, consistent with other works in the literature that use such a regularization term. Engineering efforts on this parameter, such as a temperature function to automatically increase this weight during training, may prove beneficial if the goal is to reach the maximum possible accuracy.

\textbf{Large scale impact and future directions:} The reconstruction sub-network of \textit{DeformCaps} could possibly be trained to produce a fast single-shot instance segmentation framework. At test, potentially detected objects could have their instantiation vectors reconstructed into objects' masks, then these masks would simply be resized to the predicted bounding-boxes, similar to~\cite{he2017mask} but without needing to have the initial reshape and ROI alignment required in their two-stage approach. Such an approach would likely need a rigorous balancing of the relative objective functions, as well as careful consideration of the length of the instantiation vectors relative to the complexity of the data being modeled.

Our goal is to make it computationally feasible to scale capsules up to large-scale tasks such as detection for the first time, a problem which has significantly held back research on capsule networks. It is our hope that utilizing Deformable SplitCaps with SE-Routing, deeper investigations into unique capsule properties on a wide range of large scale tasks can follow.

The development of DeformCaps and its associated algorithms, SplitCaps and SE-Routing, represents a significant advancement in the field of object detection. The potential impact at large-scale of this research can be summarized as: (1) improved object detection accuracy via reduced false positives and generalization to unusual poses/viewpoints, (2) stronger internal representation, and (3) interpretability. Future works derived from this work can be summarized as follows:

\textbf{Computational Efficiency Optimization:} While DeformCaps address the limitations of previous capsule networks, further research could explore additional methods to improve their computational efficiency, making them even more suitable for deployment on resource-constrained hardware.

\textbf{Integration with Other Architectures:} The DeformCaps framework could be explored for integration with other object detection architectures, potentially leading to even more robust and accurate detection systems.

\textbf{Scalability to Other Vision Tasks:} The success of DeformCaps in object detection opens doors for exploring their applicability to other computer vision tasks such as image segmentation, image classification at larger scales, and even video analysis.

\textbf{Theoretical Analysis:} Further research could delve deeper into the theoretical underpinnings of DeformCaps, providing a more comprehensive understanding of their strengths and limitations compared to existing object detection models.

\textbf{Acknowledgements} \par %delete if not applicable))
This project is supported by the NIH funding: R01-CA246704, R01-CA240639, U01 DK127384-02S1, and U01-CA268808.

\bibliographystyle{unsrtnat}
\bibliography{iclr2021_conference}

\end{document}